\documentclass[fleqn]{article}
\usepackage[final]{nips_2018}
\usepackage[utf8]{inputenc} 
\usepackage[T1]{fontenc}    
\usepackage{hyperref}       
\usepackage{url}            
\usepackage{booktabs}       
\usepackage{amsfonts}       
\usepackage{nicefrac}       
\usepackage{microtype}      
\usepackage{graphicx}
\usepackage{amsmath}

\title{TorchProteinLibrary: A computationally efficient, differentiable representation of protein structure}
\author{
    Georgy~Derevyanko\thanks{https://github.com/lupoglaz} \\
    Department of Chemistry and Biochemistry\\
    Concordia University\\
    Montreal, QC \\
    \texttt{georgy.derevyanko@gmail.com} \\
    \And
    Guillaume~Lamoureux\thanks{On leave from Department of Chemistry and Biochemistry, Concordia University, Montreal, QC} \\
    Department of Chemistry \\
    Center for Computational and Integrative Biology \\
    Rutgers University -- Camden \\
    Camden, NJ \\
    \texttt{guillaume.lamoureux@rutgers.edu} \\
}

\begin{document}

\maketitle
\begin{abstract}
Predicting the structure of a protein from its sequence is a cornerstone task of molecular biology. Established methods in the field, such as homology modeling and fragment assembly, appeared to have reached their limit. However, this year saw the emergence of promising new approaches: end-to-end protein structure and dynamics models, as well as reinforcement learning applied to protein folding. For these approaches to be investigated on a larger scale, an efficient implementation of their key computational primitives is required. In this paper we present a library of differentiable mappings from two standard dihedral-angle representations of protein structure (full-atom representation ``$\phi$,$\psi$,$\omega$,$\chi$'' and backbone-only representation ``$\phi$,$\psi$,$\omega$'') to atomic Cartesian coordinates. The source code and documentation can be found at \url{https://github.com/lupoglaz/TorchProteinLibrary}.
\end{abstract}

\section{Introduction}
To understand the molecular details of how any living system functions, one inevitably has to know the three-dimensional structure of a large number of proteins. Despite the recent progresses in cryo-electron microscopy~\citep{fernandez2016unravelling}, experimental approaches to this problem are unlikely to scale up, and there is a dire need for innovation in computational methods.

State-of-the-art methods that attempt to solve the protein folding problem~\citep{dill2012} usually rely on complex workflows that consist of multiple loosely interconnected operations~\citep{yang2015tasser, raman2009structure}. For the majority of these methods, a first step consists in generating a ``rough'' protein structure, using either homology modelling or some fragment-based assembly approach. The second step consists in refining this structure using one of many optimization techniques. The parameters of these two steps are usually tuned separately.

However, new computational approaches to protein folding have recently emerged, which make use of end-to-end learning. The work of AlQuraishi~\citep{Mohammed:2018} attempts to learn the positions of protein backbone atoms using a BiLSTM that predicts distributions of dihedral angles and, using a differentiable transformation, that converts these internal coordinates into atomic Cartesian coordinates. Other work under review~\citep{anonymous2019learning} tries to learn the force field parameters used by a simulation of the protein folding process, using the same differentiable conversion between internal coordinates and atomic coordinates.

We anticipate that end-to-end models will become extremely important for structural biology, due to the growing amounts of data for both protein sequences and protein structures, and the necessity of relating the two. Since the transformation between internal coordinates and atomic positions is the only known unambiguous differentiable mapping between protein sequence and protein structure, a fast implementation of this transformation is the key building block for any ``sequence-to-structure'', end-to-end model.

In this work we present \textsc{TorchProteinLibrary}, a library that implements the conversion between internal protein coordinates and atomic positions for ``full-atom'' and ``backbone-only'' models of protein structure. It also contains an implementation of the least root-mean-square deviation (LRMSD), a measure of distance in the space of protein structures that respects translational and rotational invariance.

\section{Full-atom model}

\begin{figure}
  \centering
  \includegraphics[width=0.6\columnwidth]{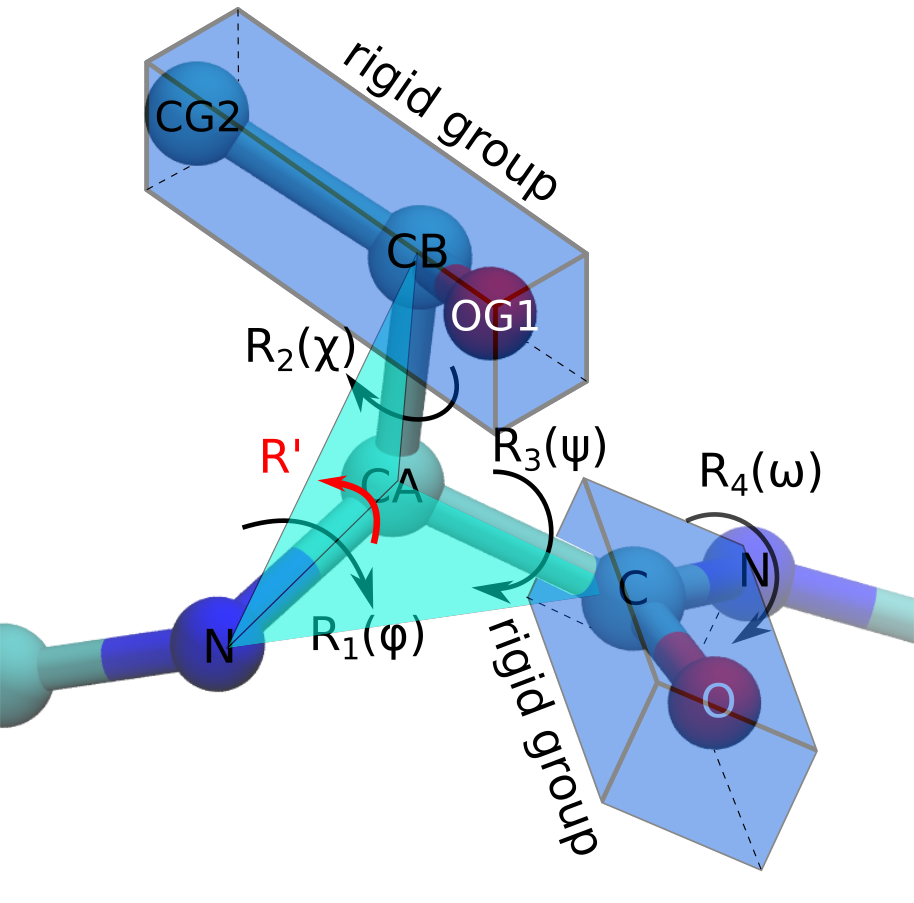}
  \caption{Example of parameterization of threonine in terms of dihedral angles $\phi$, $\psi$, $\omega$, and $\chi$. In the current implementation $\omega$ is fixed to $\pi$, which makes $R_4$ constant. Blue boxes are aligned to local coordinate systems, in which initial coordinates of rigid groups are defined. $R'$ is an out-of-plane transform that does not contain differentiable parameters.}
  \label{Fig:aminoacid}
\end{figure}

The ``full-atom'' representation of protein structure specifies the positions of all non-hydrogen atoms. (Following the usual convention, hydrogen atoms are omitted from the representation because their positions are easy to infer from the rest of the molecular structure.) The layer computes the Cartesian coordinates by building a graph of transforms, acting on standard coordinates of rigid groups of atoms. In that standard reference frame, the first atom of the rigid group is at the origin and any additional atom is at a position consistent with the stereochemistry of the group. The smallest rigid group consists of a single atom at the origin. Each amino acid conformation is described by a list of up to 7 dihedral angles. For example, Figure~\ref{Fig:aminoacid} shows a schematic representation of amino acid threonine, which has 5 rigid groups in total and is parameterized by 4 transforms.

Each transform $R_i$ is parameterized by a dihedral angle $\alpha_i$ and has the form 
\begin{equation*}
R_i(\alpha_i) = R(\alpha_i, \theta_i, d_i) = R_y (\theta_i) T_x(d_i) R_x(\alpha_i)
\end{equation*}
where $R_x(\alpha_i)$ and $R_y(\theta_i)$ are the $4\times 4$ rotation matrices about axes $x$ and $y$, respectively, and $T_x(d_i)$ is the $4\times 4$ translation matrix along axis $x$. $\theta_i$ and $d_i$ are fixed parameters for which we do not compute derivatives. They depend only on the type of the amino acid and its stereochemical properties. For instance, the first transform of the threonine parameterization of Figure~\ref{Fig:aminoacid} can be written as
\begin{equation*}
R_1(\phi) = R(\phi, \theta_1, d_1)
\end{equation*}  
where $\phi$ is the first dihedral angle (variable), $\theta_1$ is the C-N-CA angle (fixed), and $d_1$ is the N-CA bond length (also fixed).


We compute the position of every atom in a rigid group by transforming its position in the original reference frame with the appropriate matrix. In short, to get the cumulative transform of a node $M_{i}$, we take the cumulative transform $M_{\mathrm{parent}(i)}$ and multiply it by the transform of the current node $R_{i}$. In the threonine example, the cumulative transforms are as follows:
\begin{eqnarray*}
M_1 &=& M_0 R_1(\phi)\\
M_2 &=& M_0 R_1(\phi) R' R_2(\chi)\\
M_3 &=& M_0 R_1(\phi) R_3(\psi)\\
M_4 &=& M_0 R_1(\phi) R_3(\psi) R_4(\omega)
\end{eqnarray*}
$M_0$ represents the cumulative transform leading to the threonine residue considered, due to all residues ahead in the sequence. For L-amino acids, $R'$ corresponds to a counterclockwise rotation of $122.686^\circ$ about the $x$ axis, needed to properly orient the side chain.
The atomic coordinates ``$\mathbf{r}$'' are obtained by transforming the standard coordinates ``$\mathbf{r}^\circ$'' of each rigid group $i$ by its corresponding cumulative transform $M_i$. For instance, the atomic positions of the threonine residue can be written as follows:
\begin{eqnarray*}
\mathbf{r}_\mathrm{CA} &=& M_1 \mathbf{r}_\mathrm{CA}^\circ = M_1 \mathbf{0}\\
\mathbf{r}_\mathrm{CB} &=& M_2 \mathbf{r}_\mathrm{CB}^\circ = M_2 \mathbf{0}\\
\mathbf{r}_\mathrm{OG1} &=& M_2 \mathbf{r}_\mathrm{OG1}^\circ\\
\mathbf{r}_\mathrm{CG2} &=& M_2 \mathbf{r}_\mathrm{CG2}^\circ\\
\mathbf{r}_\mathrm{C} &=& M_3 \mathbf{r}_\mathrm{C}^\circ = M_3 \mathbf{0}\\
\mathbf{r}_\mathrm{O} &=& M_3 \mathbf{r}_\mathrm{O}^\circ\\
\mathbf{r}_\mathrm{N} &=& M_4 \mathbf{r}_\mathrm{N}^\circ = M_4 \mathbf{0}
\end{eqnarray*}
where $\mathbf{r}_t^\circ = (x_t^\circ, y_t^\circ, z_t^\circ, 1)^\mathrm{T}$ is the 4-component vector representing the position of atom of type $t$ in the standard reference frame and $\mathbf{0} = (0,0,0,1)^\mathrm{T}$ is the 4-component vector representing the origin.

\begin{figure}
  \centering
  \includegraphics{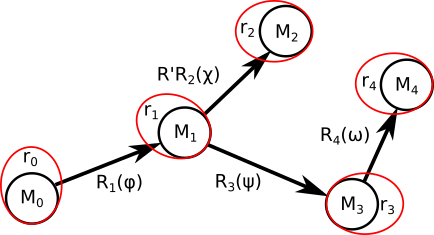}
  \caption{Example of the threonine molecular graph. Each node in the graph contains the cumulative transform $M_i$, computed during the forward pass. Rigid groups coordinates associated with each node are denoted as $\mathbf{r}_i$.}
  \label{Fig:molgraph_simple}
\end{figure}

We describe the algorithm of computing the gradients of atomic positions with respect to any dihedral angle $\alpha_i$ by considering the transformation graph from Figure~\ref{Fig:molgraph_simple}, corresponding to threonine. The graph has four nodes that contain the cumulative transformation matrices $M_1$ to $M_4$. To simplify the notation, we assume that each rigid group contains a single atom at position $\mathbf{r}_i$. The corresponding atoms in threonine would be $\mathbf{r}_1 = \mathbf{r}_\mathrm{CA}$, $\mathbf{r}_2 = \mathbf{r}_\mathrm{CB}$, $\mathbf{r}_3 = \mathbf{r}_\mathrm{C}$, and $\mathbf{r}_4 = \mathbf{r}_\mathrm{N}$. Supposing we have a function $L$ that depends only on the coordinates of these four atoms, we can write its derivative with respect to dihedral angle $\phi$ as:
\begin{equation*}
\frac{\partial L}{\partial \phi} = 
\sum_{k=1}^4 \frac{\partial L}{\partial \mathbf{r}_k}\cdot \frac{\partial \mathbf{r}_k}{\partial \phi}
\end{equation*}
where ``$\cdot$'' denotes the scalar product of two vectors.
The derivative of the first position with respect to $\phi$ can be written as:
\begin{equation*}
\frac{\partial \mathbf{r}_1}{\partial \phi} = 
M_0 \frac{\partial R_1}{\partial \phi} \mathbf{r}_1^\circ
\end{equation*}
where $\mathbf{r}_1^\circ$ represents the position of the atom in the standard reference frame. The other three derivatives can be written as:
\begin{equation*}
\frac{\partial \mathbf{r}_2}{\partial \phi} = 
M_0 \frac{\partial R_1}{\partial \phi} R' R_2 \mathbf{r}_2^\circ
,\qquad
\frac{\partial \mathbf{r}_3}{\partial \phi} = 
M_0 \frac{\partial R_1}{\partial \phi} R_3 \mathbf{r}_3^\circ
\qquad\mathrm{and}\qquad
\frac{\partial \mathbf{r}_4}{\partial \phi} = 
M_0 \frac{\partial R_1}{\partial \phi} R_3 R_4 \mathbf{r}_4^\circ
\end{equation*}
We can write those expressions using $\mathbf{r}_1$, $\mathbf{r}_2$, $\mathbf{r}_3$ and $\mathbf{r}_4$, the atomic coordinates calculated during the forward pass:
\begin{equation*}
\frac{\partial \mathbf{r}_1}{\partial \phi} = 
M_0 \frac{\partial R_1}{\partial \phi} M_1^{-1} M_1 \mathbf{r}_1^\circ =
M_0 \frac{\partial R_1}{\partial \phi} M^{-1}_1 \mathbf{r}_1
\end{equation*}
\begin{equation*}
\frac{\partial \mathbf{r}_2}{\partial \phi} = 
M_0 \frac{\partial R_1}{\partial \phi} M_1^{-1} M_1 R' R_2 \mathbf{r}_2^\circ = 
M_0 \frac{\partial R_1}{\partial \phi} M_1^{-1} M_2 \mathbf{r}_2^\circ = 
M_0 \frac{\partial R_1}{\partial \phi} M_1^{-1} \mathbf{r}_2
\end{equation*}
\begin{equation*}
\frac{\partial \mathbf{r}_3}{\partial \phi} = 
M_0 \frac{\partial R_1}{\partial \phi} M_1^{-1} M_1 R_3 \mathbf{r}_3^\circ = 
M_0 \frac{\partial R_1}{\partial \phi} M_1^{-1} M_3 \mathbf{r}_3^\circ = 
M_0 \frac{\partial R_1}{\partial \phi} M^{-1}_1 \mathbf{r}_3
\end{equation*}
\begin{equation*}
\frac{\partial \mathbf{r}_4}{\partial \phi} = 
M_0 \frac{\partial R_1}{\partial \phi} M_1^{-1} M_1 R_3 R_4 \mathbf{r}_4^\circ = 
M_0 \frac{\partial R_1}{\partial \phi} M_1^{-1} M_4 \mathbf{r}_4^\circ = 
M_0 \frac{\partial R_1}{\partial \phi} M^{-1}_1 \mathbf{r}_4
\end{equation*}
The expression for the derivative becomes:
\begin{equation*}
\frac{\partial L}{\partial \phi} = 
\sum_{k=1}^4 \frac{\partial L}{\partial \mathbf{r}_k}\cdot M_0 \frac{\partial R_1}{\partial \phi} M^{-1}_1 \mathbf{r}_k
\end{equation*}
In this formula index $k$ iterates over the children of node 0, and matrix $M_0 \frac{\partial R_1}{\partial \phi} M^{-1}_1$ can be computed efficiently. This expression can be generalized to any graph without loops. For $L$ a function of all atomic positions, the derivative with respect to any dihedral angle $\theta_i$ of the node $i$ is:
\begin{equation}
\label{Eq:FAMBackward}
\frac{\partial L}{\partial \theta_i} = 
\sum_{k \in \mathrm{children}(i)} 
\frac{\partial L}{\partial \mathbf{r}_k}\cdot 
M_{i} \frac{\partial R_i}{\partial \theta_i} M^{-1}_{i+1} 
\mathbf{r}_k
\end{equation}
This sum is computed during a backward depth-first propagation through the graph. Matrices $F_i = M_{i} \frac{\partial R_i}{\partial \theta_i} M^{-1}_{i+1}$, however, are computed during the forward pass. The library presented in this work implements the forward and backward passes on CPU.

\section{Backbone model}

\begin{figure}
\centering
\includegraphics[width=0.6\columnwidth]{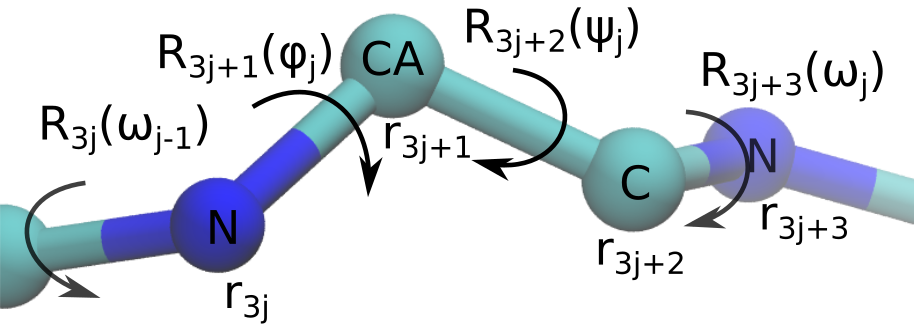}
\caption{Illustration of parameterization of one amino acid number $j$ in the backbone model.}
\label{Fig:reducedmodel}
\end{figure}

Another widely used protein representation is the ``backbone'' model, shown on Figure~\ref{Fig:reducedmodel}, for which we compute only three atomic positions per residue (for CA, C, and N atoms). The backbone O atoms are omitted but their positions can be easily inferred from the positions of the other three atoms. In this reduced representation, the amino acid side chains are ignored.

The backbone model, unlike the full-atom model, can be efficiently implemented on GPU. The key to efficient parallel implementation is that $\partial \mathbf{r}_i/\partial \theta_j$, the derivatives of the coordinates with respect to parameters of the model, can be computed independently of one another. Here we describe the detailed computation of the coordinates and write down the derivatives in terms of quantities saved during the forward pass.

The position of the $i$-th atom in the chain is:
\begin{equation*}
\mathbf{r}_i = R_0 R_1 \cdots R_i \mathbf{0}
\end{equation*}
where $\mathbf{0} = (0,0,0,1)^\mathrm{T}$ and $R_i$ are transformation matrices parameterized by dihedral angle $\alpha$:
\begin{equation*}
R(\alpha, \theta, d) = \begin{bmatrix}
\cos(\theta) & \sin(\alpha)\sin(\theta) & \cos(\alpha)\sin(\theta) & d\cos(\theta) \\
0 & \cos(\alpha) & -\sin(\alpha) & 0 \\
-\sin(\theta) & \sin(\alpha)\cos(\theta) & \cos(\alpha)\cos(\theta) & -d\sin(\theta) \\
0 & 0 & 0 & 1
\end{bmatrix}
\end{equation*}

The sequence of transformations is defined for amino acid sequence indexed with $j\in [0,L)$, where $L$ is the length of the sequence. If $j<0$, then the transformation is the identity matrix, otherwise:
\begin{itemize}
    \item C-N peptide bond of residue $j-1$:\\
		  Atomic index: $i = 3j$\\
		  Transformation: $ R_i = R(\omega_j, \pi - 2.1186, 1.330)$
	\item N-CA bond of residue $j$:\\
		  Atomic index: $i = 3j + 1$\\
		  Transformation: $ R_i = R(\phi_{j}, \pi - 1.9391, 1.460)$
	\item CA-C bond of residue $j$:\\
		  Atomic index: $i = 3j + 2$\\
		  Transformation: $ R_i = R(\psi_{j}, \pi - 2.0610, 1.525)$
    
\end{itemize}

In the current implementation the angle $\omega_{j-1}$ is fixed to $\pi$, corresponding to the \emph{trans} conformation of the peptide bond. During the forward pass, we save the cumulative transformation matrices for each atom:
\begin{equation*}
M_i = R_0 R_1 \cdots R_i
\end{equation*}
Notice that atom $3j$ is always N, atom $3j + 1$ is always CA, and atom $3j+2$ is always C. Thus
transformation matrices $R_{3j+1}$ and $R_{3j+2}$ depend on the angles $\phi_j$ and $\psi_j$, respectively. During the backward pass, we first compute the gradient of $\mathbf{r}_i$ with respect to each ``$\phi$'' and ``$\psi$'' angle:
\begin{equation*}
\frac{\partial \mathbf{r}_i}{\partial \phi_j} = R_0 R_1 \cdots \frac{\partial R_{3j+1}}{\partial \phi_j} \cdots R_i \mathbf{0}
\qquad \mathrm{and} \qquad
\frac{\partial \mathbf{r}_i}{\partial \psi_j} = R_0 R_1 \cdots \frac{\partial R_{3j+2}}{\partial \psi_j} \cdots R_i \mathbf{0}
\end{equation*}
We can rewrite these expressions using the matrices $M$, saved during the forward pass:
\begin{equation*}
\frac{\partial \mathbf{r}_i}{\partial \phi_j} = M_{3j}  \frac{\partial R_{3j+1}}{\partial \phi_j} M_{3j+1}^{-1} M_{i} \mathbf{0}
\qquad \mathrm{and} \qquad
\frac{\partial \mathbf{r}_i}{\partial \psi_j} = M_{3j+1}  \frac{\partial R_{3j+2}}{\partial \psi_j} M_{3j+2}^{-1} M_{i} \mathbf{0}
\end{equation*}

The inverses of matrices $M_k$ have simple forms and can be computed on the fly during the backward pass. This allows to compute all derivatives simultaneously on GPU. To compute the derivatives of function $L$, which depends on the derivatives of the atomic coordinates with respect to the input angles, we have to calculate the following sums:
\begin{equation}
\label{Eq:BackboneGradOutput}
\sum_i \frac{\partial L}{\partial \mathbf{r}_i}\cdot \frac{\partial \mathbf{r}_i}{\partial \phi_j}
\qquad \mathrm{and} \qquad
\sum_i \frac{\partial L}{\partial \mathbf{r}_i}\cdot \frac{\partial \mathbf{r}_i}{\partial \psi_j}
\end{equation}
which can be efficiently computed on GPU.

\section{Loss function}

The training of any protein structure prediction model requires some measure of how close two structures of the same sequence are. While there are multiple variations of such a measure, the most common one is the least root-mean-square deviation (LRMSD) of atomic Cartesian coordinates:
\begin{equation*}
\mathrm{LRMSD} = \min_M \sqrt{ \sum_i^{N_\mathrm{atoms}}{\frac{|\mathbf{x}_i - M\mathbf{y}_i|^2}{N_\mathrm{atoms}} }}
\end{equation*}
Here, $\mathbf{x}_i$ and $\mathbf{y}_i$ are the atomic positions of the ``target'' and ``input'' structure, respectively, and the root-mean-square deviation is minimized over all possible rigid transformation matrices $M$. This measure is invariant with respect to rotations and translations of the two structures being compared.

\textsc{TorchProteinLibrary} contains an implementation of the algorithm of Coutsias, Seok and Dill~\citep{coutsias2004using}. This algorithm computes LRMSD and its derivative with respect to the coordinates of one of the structures without explicit minimization. We briefly outline the key steps of the algorithm but a detailed derivation can be found in Ref.~\citep{coutsias2004using}.

We first move both target and input structures (positions $\mathbf{x}_i$ and $\mathbf{y}_i$) so that their barycenters are at the origin, then we compute the correlation matrix $R$:
\begin{equation*}
R = \sum_i^{N_\mathrm{atoms}} \mathbf{x}_i \mathbf{y}^\mathrm{T}_i
\end{equation*}
Using this $3\times 3$ matrix we compute the following $4\times 4$ matrix $T$:
\begin{equation*}
T = \begin{bmatrix}
R_{11} + R_{22} + R_{33} & R_{23} - R_{32} & R_{31} - R_{13} & R_{12} - R_{21} \\
R_{23} - R_{32} & R_{11} - R_{22} - R_{33} & R_{12} + R_{21} & R_{13} + R_{31} \\
R_{31} - R_{13} & R_{12} + R_{21} & -R_{11} + R_{22} - R_{33} & R_{23} + R_{32} \\
R_{12} - R_{21} & R_{13} + R_{31} & R_{23} + R_{32} & -R_{11} - R_{22} + R_{33} \\
\end{bmatrix}
\end{equation*}
We then compute $\lambda$, the maximum eigenvalue of matrix $T$, and its associated eigenvector $\mathbf{q}$. This eigenvector corresponds to the quaternion that gives the optimal rotation of one structure with respect to the other. The rotation matrix can be computed as follows:
\begin{equation*}
U = \begin{bmatrix}
q^2_0 + q^2_1 - q^2_2 - q^2_3 & 2(q_1 q_2 - q_0 q_3) & 2(q_1 q_3 + q_0 q_2) \\
2(q_1 q_2 + q_0 q_3) & q^2_0 - q^2_1 + q^2_2 - q^2_3 & 2(q_2 q_3 - q_0 q_1) \\
2(q_1 q_3 - q_0 q_2) & 2(q_2 q_3 + q_0 q_1) & q^2_0 - q^2_1 - q^2_2 + q^2_3
\end{bmatrix}
\end{equation*}
The LRMSD is computed using the formula:
\begin{equation*}
\mathrm{LRMSD} = \sqrt{ \frac{ \sum_i^{N_\mathrm{atoms}} \left( |\mathbf{x}_i|^2 + |\mathbf{y}_i|^2 \right) - 2\lambda }{N_\mathrm{atoms}}}
\end{equation*}
The derivative of LRMSD with respect to the input coordinates is computed using the formula:
\begin{equation*}
\frac{\partial \mathrm{LRMSD}}{\partial \mathbf{x}_i} = \mathbf{x}_i - U^\mathrm{T} \mathbf{y}_i
\end{equation*}
This expression, combined with Eqs.~\ref{Eq:FAMBackward} or \ref{Eq:BackboneGradOutput}, allows any sequence-based model predicting internal coordinates of proteins to be directly trained on known protein structures, using LRMSD as a loss function.

\section{Benchmarks}

Here we give estimates of the run times of the modules described above and a simple baseline for comparison. We perform the measurements using a Titan X Maxwell GPU with Intel Core i7-5930K CPU machine and PyTorch version 0.4.1 CUDA 9.2 build.

\begin{figure}
  \centering
  \includegraphics[width=0.7\columnwidth]{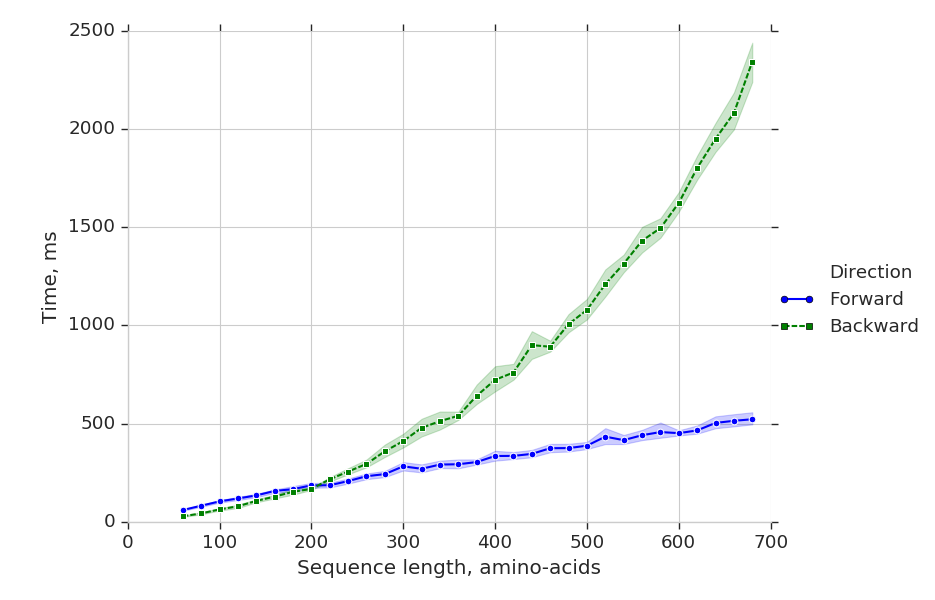}
  \caption{Scaling of computation time of forward and backward passes for the full-atom model. The batch size was set to 32 and amino acid sequences were generated at random for each measurement. We performed 10 measurements per data point and plotted the 95\% confidence interval.}
  \label{Fig:FAMTime}
\end{figure}

Figure~\ref{Fig:FAMTime} shows the scaling of computation time of the forward and backward passes for the full-atom model. We see that the computational complexity of the backward pass is $O(L^2)$, where $L$ is the sequence length. The reason for this quadratic scaling is that we compute Eq.~\ref{Eq:FAMBackward} using depth-first graph traversal. In principle, this layer can be further optimized by unfolding the graph and computing the gradients simultaneously on GPU. (We are planning to incorporate this optimization in the next version of the library.)

\begin{figure}
  \centering
  \includegraphics[width=0.7\columnwidth]{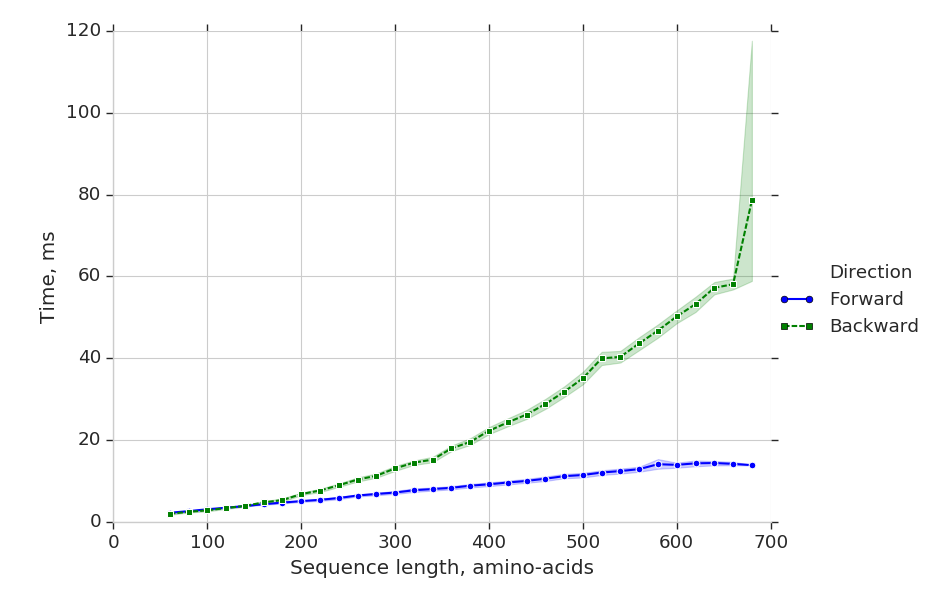}
  \caption{Scaling of computation time for forward and backward passes for the backbone model. The batch size was set to 32. We performed 10 measurements per data point and plotted the 95\% confidence interval.}
  \label{Fig:BackboneTime}
\end{figure}

Figure \ref{Fig:BackboneTime} shows the computation time scaling for the backbone protein model with the growth of the sequence length. While the computational complexity for the backward pass still scales like $O(L^2)$, the scaling coefficient is smaller than for the full-atom model. Here, the presence of quadratic scaling can be attributed to GPU hardware limitations. However, we expect this scaling coefficient to decrease with an increase in the number of CUDA cores, due to increasingly efficient parallelization. Another contributing factor to this scaling behavior is the computation of the sums in Eq.~\ref{Eq:BackboneGradOutput}, which currently does not rely on the ``reduce'' algorithm for parallelization.

\begin{figure}
  \centering
  \includegraphics[width=0.7\columnwidth]{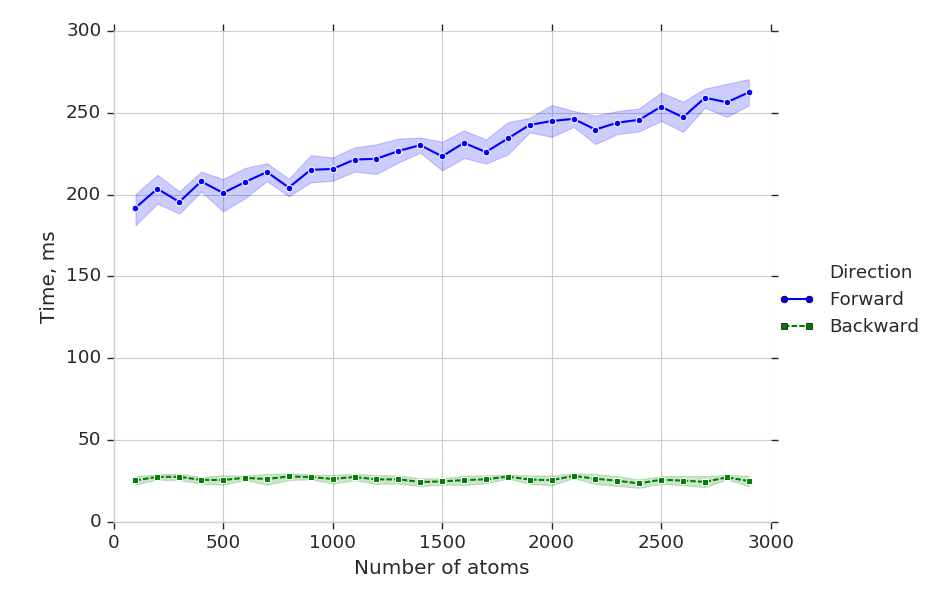}
  \caption{Scaling of computation time for forward and backward passes of LRMSD. The batch size was set to 32. We performed 10 measurements per data point and plotted the 95\% confidence interval.}
  \label{Fig:RMSDTime}
\end{figure}

Finally, the LRMSD layer computation time is shown on Figure~\ref{Fig:RMSDTime}. The forward pass run time exhibits the expected linear behavior.

\begin{figure}
  \centering
  \includegraphics[width=0.7\columnwidth]{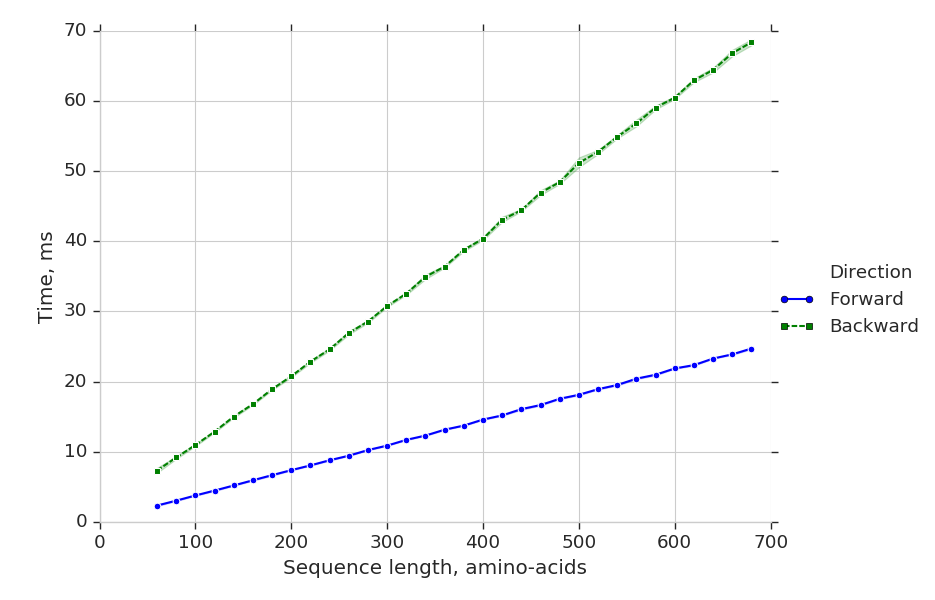}
  \caption{Scaling of computation time for forward and backward passes of a one-layer LSTM on GPU. The batch size was set to 32, the number of input features was 128, and the number of hidden units was set to 256. We performed 10 measurements per data point and plotted the 95\% confidence interval.}
  \label{Fig:LSTMTime}
\end{figure}

To have a meaningful reference timescale for the computational times of the layers implemented in the library, we measured forward and backward computation times of an LSTM model~\citep{hochreiter1997long} on GPU, as implemented in PyTorch~\citep{paszke2017automatic}. The batch size of the input is 32 and the number of input features is 128. The LSTM has 256 hidden units and one layer. Figure~\ref{Fig:LSTMTime} shows the scaling of the forward and backward passes of this model with the growth of the input length. We see that the LSTM and backbone models have comparable computation times.

\begin{figure}
  \centering
  \includegraphics[width=0.7\columnwidth]{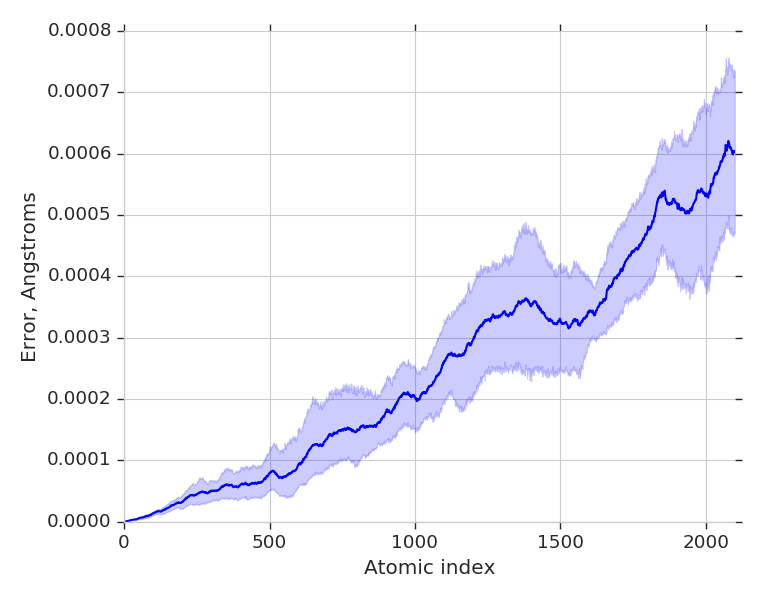}
  \caption{Scaling of computation error for forward pass for the backbone model. Atomic index corresponds to the atom numbers in backbone protein model, for example index 2100 corresponds to the CA in residue number 700. We performed 10 measurements per data point and plotted the 95\% confidence interval.}
  \label{Fig:ErrorEstimate}
\end{figure}

Another important concern regarding the backbone model is the numerical stability of chained matrix multiplications using single-precision arithmetic during the forward pass. To estimate the error of the computation we compared the output from a forward pass of the backbone protein model to that of the full-atom model, implemented using double-precision arithmetic on CPU. We generate random input angles for the backbone of a protein and pass them through both full-atom and backbone layers, then compute the distances between equivalent backbone atoms in both models. Figure~\ref{Fig:ErrorEstimate} shows that the resulting error is negligible for proteins sequences of any (realistic) length.

\section{Conclusion}

In this paper we have described two differentiable representations that allow the mapping of protein sequences to protein atomic coordinates. These representations enable the development of end-to-end differentiable models and of model-based reinforcement learning~\citep{berkenkamp2017safe}. We believe these new approaches hold great promise for protein folding and protein modelling in general.

It should be mentioned that the scope of \textsc{TorchProteinLibrary} is not limited to these layers only. To gain access to the chemical properties of protein molecules, functions mapping atomic coordinates to a scalar value have to be defined. For that purpose, we have also implemented a representation of the atomic coordinates as a density map on a three-dimensional (3D) grid. This representation can then be transformed into a scalar using 3D convolutional networks. This particular 3D representation allows to circumvent some common problems associated with pairwise potentials. 
Currently available protein structure datasets are often insufficient to extract meaningful pairwise potentials for all combinations of atom types.

Another research direction for which this library is expected to be useful is the prediction of protein-protein interactions (PPIs). We have implemented differentiable volume convolutions using cuFFT~\citep{nvidia2010cufft}. These operations are at the core of most algorithms for exhaustive rigid protein-protein docking~\citep{katchalski1992molecular}. By constructing a differentiable model that maps two sets of atomic coordinates to the distribution of relative rotations and translations of one set with respect to the other, one can in principle learn to dock proteins directly from experimental data.

\section*{Acknowledgments}
We thank Yoshua Bengio for hosting one of us (G.D.) at the Montreal Institute for Learning Algorithms (MILA) during some of the critical stages of the project, and for access to the MILA computational resources.
This work was supported by a grant from the Natural Sciences and Engineering Research Council of Canada to G.L.\ (RGPIN 355789).

\bibliographystyle{authordate1}
\bibliography{citations}
\end{document}